\documentclass[runningheads]{llncs}

\usepackage{amsmath,amsfonts}

\usepackage{booktabs}
\usepackage{bm}
\usepackage{color}
\usepackage{listings}
\usepackage{siunitx}

\usepackage[linesnumbered,ruled]{algorithm2e}

\usepackage{graphicx}
\begin{document}
\title{MaxDropoutV2: An Improved Method to Drop out Neurons in Convolutional Neural Networks\thanks{The authors are grateful to FAPESP grants \#2013/07375-0, \#2014/12236-1, \#2019/07665-4, Petrobras grant \#2017/00285-6, CNPq grants \#307066/2017-7, and \#427968/2018-6, as well as the Engineering and Physical Sciences Research Council (EPSRC) grant EP/T021063/1.}}
\titlerunning{MaxDropoutV2: An Improved Method to Drop out Neurons in CNNs}
%
\author{\IEEEauthorblockN{Claudio Filipi Goncalves do Santos, Mateus Roder, Leandro A. Passos, Jo\~ao P. Papa}
\IEEEauthorblockA{Department of Computing \\
S\~ao Paulo State University\\
Bauru, Brazil \\
\{claudio.., mateus.roder, leandro.passos, joao.papa\}@unesp.br}
}

\author{Claudio Filipi Goncalves do Santos\inst{1}\orcidID{0000-1111-2222-3333} \and
Mateus Roder\inst{2}\orcidID{1111-2222-3333-4444} \and
Leandro A. Passos\inst{3}\orcidID{2222--3333-4444-5555} \and
Jo\~ao P. Papa\inst{2}\orcidID{2222--3333-4444-5555}}
\authorrunning{C. Santos et al.}
%
\institute{Federal University of S\~ao Carlos, Brazil\\
\email{cfsantos@ufscar.br}
\and
S\~ao Paulo State University, Brazil\\
\email{\{mateus.roder,joao.papa\}@unesp.br}
\and
University of Wolverhampton, United Kingdom\\
\email{l.passosjunior@wlv.ac.uk}}
\maketitle              
\begin{abstract}
In the last decade, exponential data growth supplied the machine learning-based algorithms' capacity and enabled their usage in daily life activities. Additionally, such an improvement is partially explained due to the advent of deep learning techniques, i.e., stacks of simple architectures that end up in more complex models. Although both factors produce outstanding results, they also pose drawbacks regarding the learning process since training complex models denotes an expensive task and results are prone to overfit the training data. A supervised regularization technique called MaxDropout was recently proposed to tackle the latter, providing several improvements concerning traditional regularization approaches. In this paper, we present its improved version called MaxDropoutV2. Results considering two public datasets show that the model performs faster than the standard version and, in most cases, provides more accurate results.

\end{abstract}

\section{Introduction}
\label{s.intro}

\begin{sloppypar}
The last decades witnessed a true revolution in people's daily life habits. Computer-based approaches assume the central role in this process, exerting fundamental influence in basic human tasks, such as communication and interaction, entertainment, working, studying, driving, and so on. Among such approaches, machine learning techniques, especially a subfield usually called deep learning, occupy one of the top positions of importance in this context since they empower computers with the ability to act reasonably in an autonomous fashion.
\end{sloppypar}

Deep learning regards a family of machine learning approaches that stacks an assortment of simpler models. The bottommost model's output feeds the next layer, and so on consecutively, with a set of possible intermediate operations among layers. The paradigm experienced exponential growth and magnificent popularity in the last years due to remarkable results in virtually any field of application, ranging from medicine~\cite{sun2019idiopathic,santos2021Covid,passosECCOMAS19} and biology~\cite{RoderICAISC:20} to speech recognition~\cite{noda2015audio} and computer vision~\cite{santanaIEEE-IS:19}.

Despite the success mentioned above, deep learning approaches still suffer from a drawback very commonly observed in real-world applications, i.e., the lack of sufficient data for training the model. Such a constraint affects the learning procedure in two main aspects: (i) poor classification rates or (ii) overfitting to training data. The former is usually addressed by changing to a more robust model, which generally leads to the second problem, i.e., overfitting. Regarding the latter, many works tackled the problem using regularization approaches, such as the well-known batch normalization~\cite{ioffe2015batch}, which normalizes the data traveling from one layer to the other, and dropout~\cite{srivastava2014dropout}, which randomly turns-off some neurons and forces the layer to generate sparse outputs.

Even though dropout presents itself as an elegant solution to solve overfitting issues, Santos et al.~\cite{maxdropout} claim that deactivating neurons at random may impact negatively in the learning process, slowing down the convergence. To alleviate this impact, the authors proposed the so-called MaxDropout, an alternative that considers deactivating only the most active neurons, forcing less active neurons to prosecute more intensively in the learning procedure and produce more informative features. 

MaxDropout obtained significant results considering image classification's task, however, at the cost of considerable computational cost. This paper addresses such an issue by proposing MaxDropoutV2, an improved and optimized version of MaxDropout capable of obtaining similar results with higher performance and substantial reduction of the computational burden.

Therefore, the main contributions of this work are presented as follows: 

\begin{itemize}
    \item to propose a novel regularization approach called MaxDropoutV2, which stands for an improved and optimized version of MaxDropout; 
    \item to evaluate MaxDropoutV2 overall accuracy and training time performance, comparing with the original MaxDropout and other regularization approaches; and
    \item to foster the literature regarding regularization algorithms and deep learning in general.
\end{itemize}

The remainder of this paper is organized as follows. Section~\ref{s.related} introduces the main works regarding Dropout and its variation, while Section~\ref{s.proposed} presents the proposed approach. Further, Sections~\ref{s.methodology} and~\ref{s.experiments} describe the methodology adopted in this work and the experimental results, respectively. Finally, Section~\ref{s.conclusion} states conclusions and future work.

\section{Related Works}
\label{s.related}

The employment of regularization methods for training deep neural networks (DNNs) architectures is a well-known practice, and its use is almost always considered by default. The focus of such approaches is helping DNNs to avoid or prevent overfitting problems, which reduce their generalization capability. Besides, regularization methods also allow DNNs to achieve better results considering the testing phase since the model becomes more robust to unseen data.

Batch Normalization (BN) is a well-known regularization method that employs the concept of normalizing the output of a given layer at every iteration in the training process. In its seminal work, Ioffe and Szegedy~\cite{ioffe2015batch} demonstrated that the technique is capable of speeding up the convergence regarding the task of classification. Further, several other works~\cite{zhang2017beyond,simon2016imagenet,wang2017gated} highlighted its importance, including the current state-of-the-art on image classification~\cite{tan2019efficientnet}.

Among the most commonly employed techniques for DNN regularization is the Dropout, which is usually applied to train such networks in most of the frameworks used for the task. Developed by Srivastava et al.~\cite{srivastava2014dropout}, Dropout shows significant improvements in a wide variety of applications of neural networks, like image classification, speech recognition, and more. The standard approach has a simple and efficient work procedure, in which a mask that directly multiplies the weight connections is created at training time for each batch. Such a mask follows a Bernoulli distribution, i.e., it assign values $0$ with a probability $p$ and $1$ with a probability $1 - p$. The authors showed that the best value for $p$ in hidden layers is $0.5$.  During training, the random mask varies, which means that some neurons will be deactivated while others will work normally. 

Following the initial development of the Dropout method, Wang and Manning~\cite{wang2013fast} focused on exploring different sampling strategies, considering that each batch corresponds to a new subnetwork taken into account since different units are dropped out. In this manner, the authors highlighted that the Dropout represents an approximation of a Markov chain executed several times during training time. Also, the Bernoulli distribution tends to a Normal distribution in a high dimensional space, such that Dropout performs best without sampling.

Similarly, Kingma et al.~\cite{kingma2015variational} proposed the Variational Dropout, which is a generalization of the Gaussian Dropout with the particularity of learning the dropout rate instead of randomly select one value. The authors aimed to reduce the variance of the stochastic gradients considering the variational Bayesian inference of a posterior over the model parameters, retaining the parallelization by investigating the reparametrization approach.

Further, Gal et al.~\cite{gal2017concrete} proposed a new Dropout variant to reinforcement learning models. Such a method aims to improve the performance and calibrate the uncertainties once it is an intrinsic property of the Dropout. The proposed approach allows the agent to adapt its uncertainty dynamically as more data is provided. Molchanov et al.~\cite{molchanov2017variational} explored the Variational Dropout proposed by Kingma et al.~\cite{kingma2015variational}. The authors generalized the method to situations where the dropout rates are unbounded, giving very sparse solutions in fully-connected and convolutional layers. Moreover, they achieved a reduction in the number of parameters up to $280$ times on LeNet architectures and up to $68$ times on VGG-like networks with a small decrease in accuracy rates. Such a fact highlights the importance of sparsity for robustness and parameter reduction, while the overall performance for ``simpler'' models can be improved.

Another class of regularization methods emerged in parallel, i.e., techniques that change the neural network's input. Among such methods, one can refer to the Cutout~\cite{cutout}, which works by cutting off/removing a region of the input image and setting such pixels at zero values. Such a simple approach provided relevant results in several datasets. In a similar fashion emerged the RandomErasing~\cite{zhong2020random}, which works by changing the pixel values at random for a given region in the input, instead of setting these values for zero.

Roder et al.~\cite{Roder:20EDrop} proposed the Energy-based Dropout, a method that makes conscious decisions whether a neuron should be dropped or not based on the energy analysis. The authors designed such a regularization method by correlating neurons and the model’s energy as an index of importance level for further applying it to energy-based models, as Restricted Boltzmann Machines.


\section{MaxDropoutV2 as an improved version of MaxDropout}
\label{s.proposed}

This section provides an in-depth introduction to MaxDropout-based learning.

\subsection{MaxDropout}

MaxDropout~\cite{maxdropout} is a Dropout-inspired~\cite{dropout} regularization task designed to avoid overfitting on deep learning training methods. The main difference between both techniques is that, while Dropout randomly selects a set of neurons to be cut off according to a Bernoulli distribution, MaxDropout establishes a threshold value, in which only neurons whose activation values higher than this threshold are considered in the process. Results provided in~\cite{maxdropout} show that excluding neurons using their values instead of the likelihood from a stochastic distribution while training convolutional neuron networks produces more accurate classification rates.

Algorithm~\ref{lst:maxdropout} implements the MaxDropout approach. Line $2$ generates a normalized representation of the input tensor. Line $3$ attributes the normalized value to a vector to be returned. Further, Lines $4$ and $5$ set this value to $0$ where the normalized tensor is bigger than the threshold. This process is only performed during training. Concerning the inference, the original values of the tensor are used.

\begin{lstlisting}[language=Python, numbers=left, xleftmargin=2em, caption={Original MaxDropout code}, label={lst:maxdropout}]
def MaxDropout(tensor, threshold):
    norm_tensor = normalize(tensor)
    return_tensor = norm_tensor
    if norm_tensor > threshold:
        return_tensor = 0 where
    return return_tensor 

\end{lstlisting}

Even though MaxDropout obtained satisfactory results for the task, it was not tailored-designed for Convolutional Neural Networks (CNNs), thus presenting two main drawbacks:
\begin{itemize}
\item it does not consider the feature map spacial distribution produced from a CNN layer output since it relies on individual neurons, independently of their location on a tensor; and
\item it evaluates every single neuron from a tensor, which is computationally expensive.
\end{itemize}

Such drawbacks motivated the development of an improved version of the model, namely MaxDropoutV2, which addresses the issues mentioned above and provides a faster and more effective approach. The following section describes the technique.

\subsection{MaxDropoutV2}
\label{ss.proposed}

The main difference between MaxDropout and MaxDropoutV2 is that the latter relies on a more representative feature space. While MaxDropout compares the values from each neuron directly, MaxDropoutV2 sums up these feature maps considering the depth axis, thus providing a bidimensional representation. In a nutshell, consider a CNN layer output tensor with dimensions $32\times 32\times 64$. The original MaxDropout performs $32\times 32\times 64$, i.e., $65,536$ comparisons. The proposed method sums up the values of the tensor over axis one (which would be the depth of the tensor) for each $32\times32$ kernel, thus performing only $1,024$ comparisons.

Algorithm~\ref{lst:maxdropout_v2} provides the implementation of the proposed approach. Line $2$ performs the sum in the depth axis. Similar to Algorithm~\ref{lst:maxdropout}, Line $3$ generates a normalized representation of the sum in depth of the input tensor. Line $4$ creates the mask that defines what positions of the original tensor should be dropped, i.e., set to $0$. Notice that the process is performed faster in MaxDropoutV2 due to the reduced dimensionality of the tensor. Further, in Lines $5$ and $6$, the tensor is unsqueezed and repeated so the mask can be used along all the tensor dimensions. Finally, the mask is applied to the tensor in Line $8$ and returned in Line $9$. These operations are only performed during training, similar to the original.

\begin{lstlisting}[language=Python, numbers=left, xleftmargin=2em, caption={MaxDropoutV2 code}, label={lst:maxdropout_v2}]
def MaxDropout_V2(tensor, threshold):
  sum_axis = sum tensor along axis 1 
  sum_axis = normalized(sum_axis)
  mask = 0 where sum_axis > threshold
  mask_tensor = tensor.shape[0] 
  repetitions of mask

  return_tensor = tensor * mask_tensor
  return return_tensor

\end{lstlisting}

\begin{figure}[!ht]
  \centerline{\begin{tabular}{cc}
      \includegraphics[width=0.317\columnwidth]{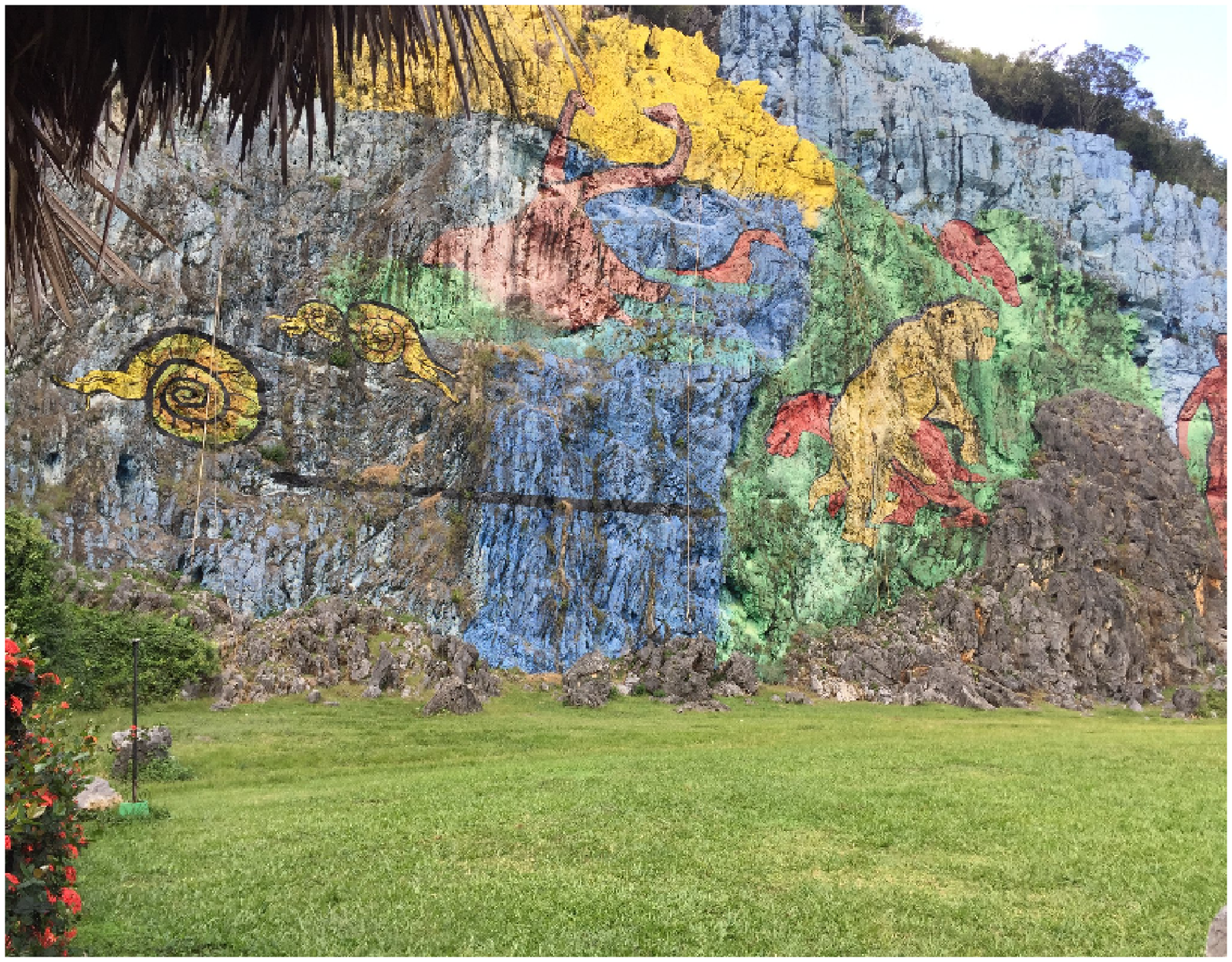} &
      \includegraphics[width=0.317\columnwidth]{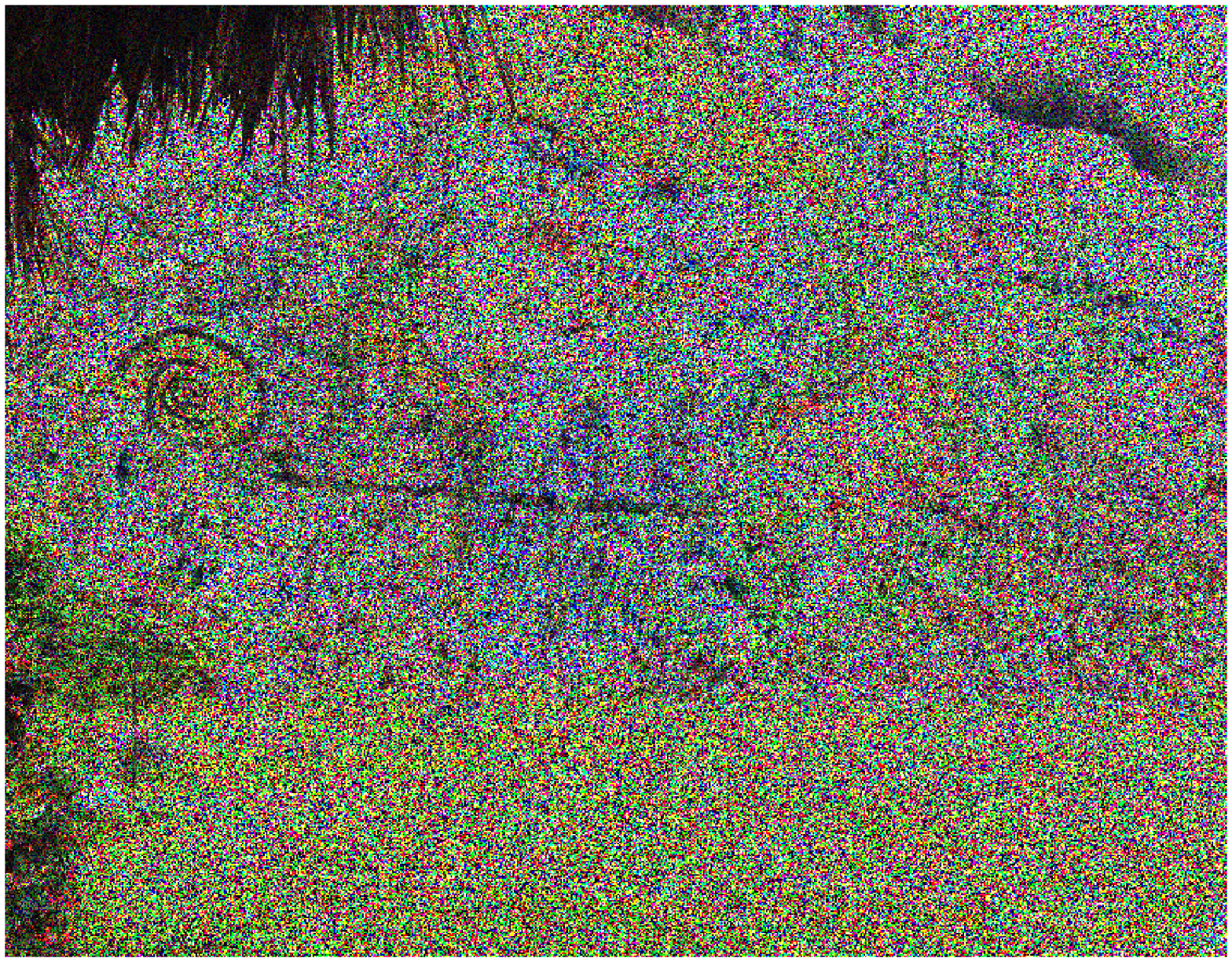} \\
      (a) & (b)\\
      \includegraphics[width=0.317\columnwidth]{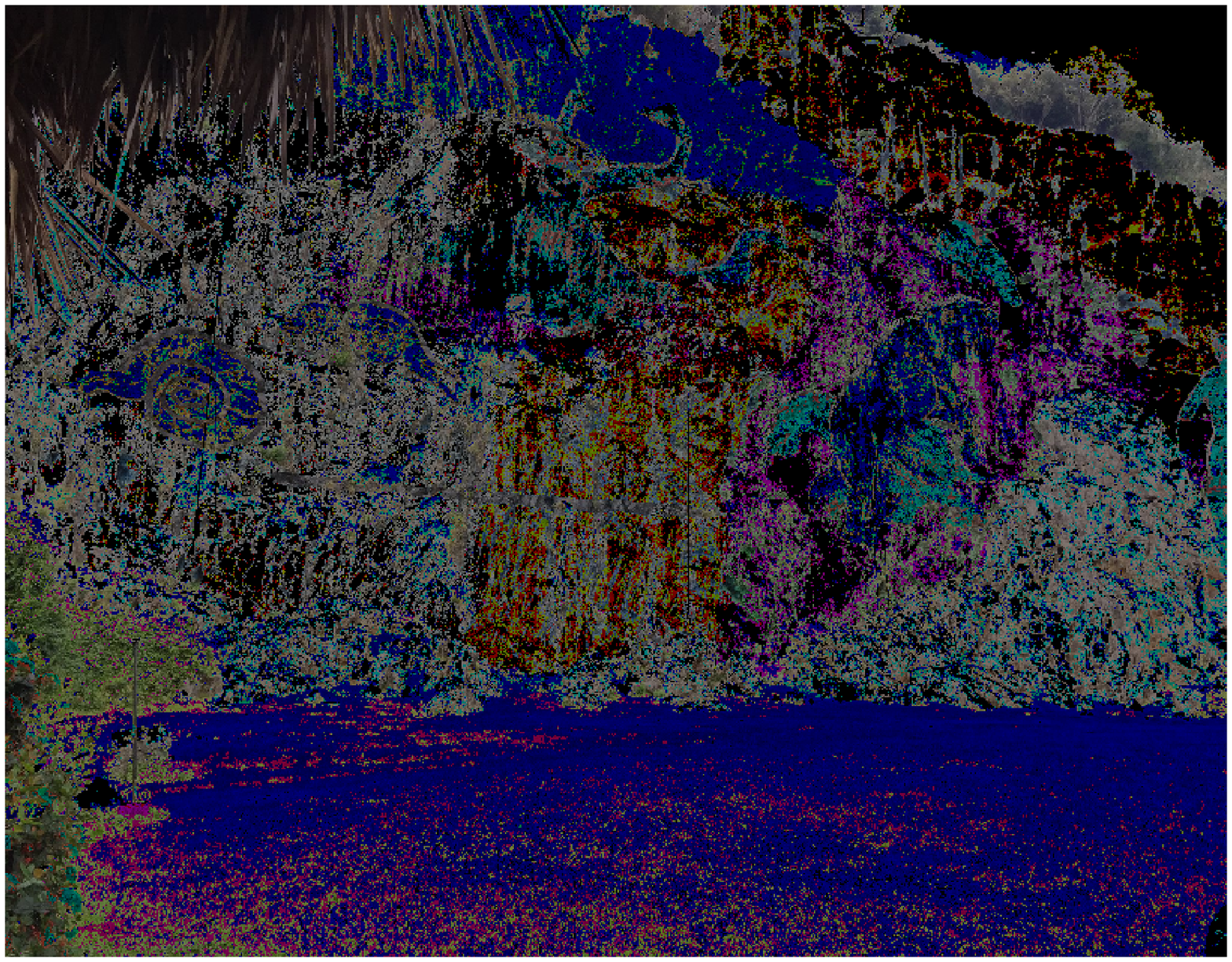} &
      \includegraphics[width=0.317\columnwidth]{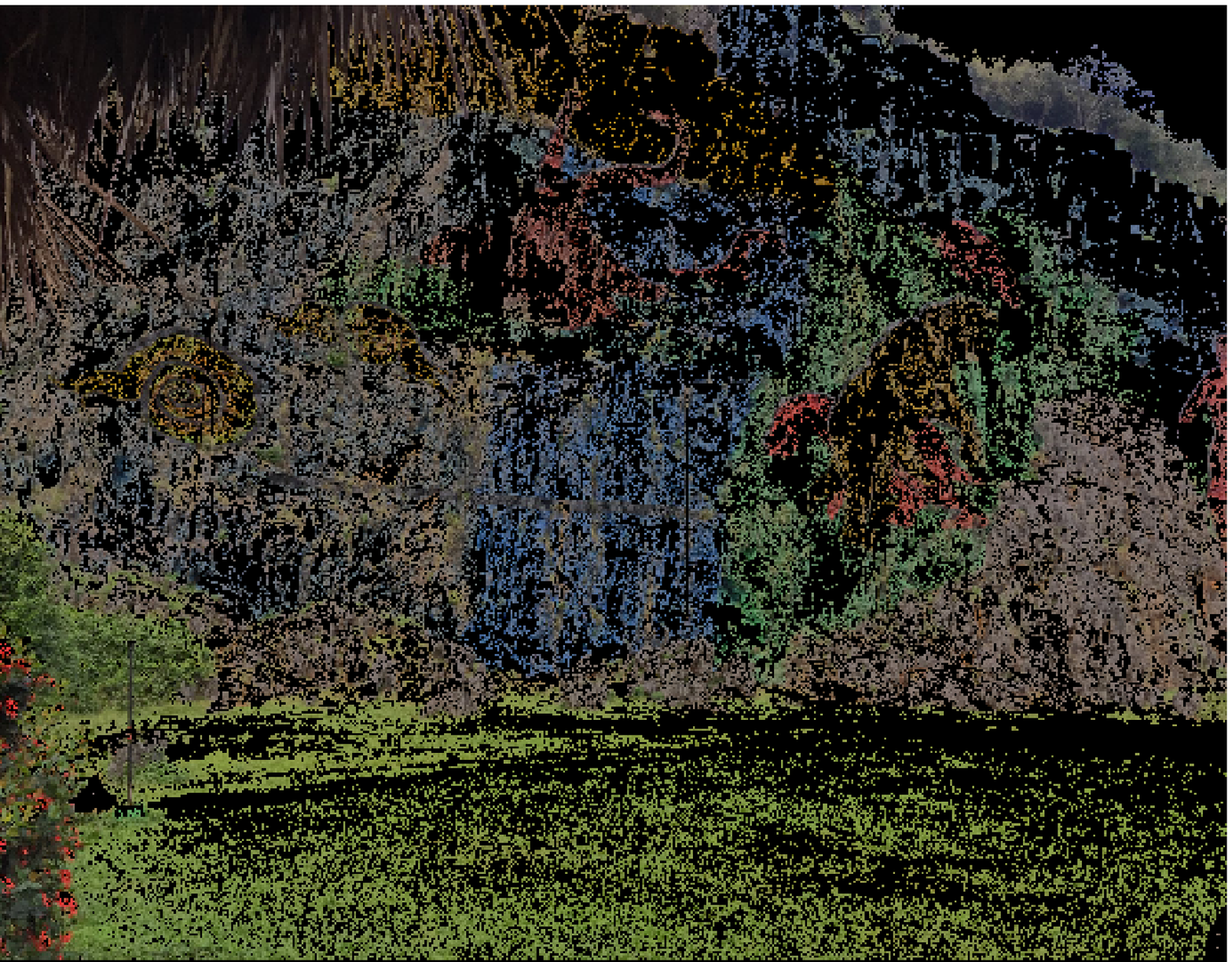} \\
      (c) & (d)
  \end{tabular}}
\caption{Simulation using colored images. The original colored image is presented in (a), and its outcomes are bestowed after (b) Dropout, (c) MaxDropout, and (d) MaxDropoutV2 transformations using a dropout rate of $50$\%.}
\label{f.simulation}
\end{figure}

Fig.~\ref{f.simulation} depicts an example of application, presenting an original image in Fig.~\ref{f.simulation}a and a simulation of output colors considering Dropout, MaxDropout, and MaxDropoutV2, for Figs.~\ref{f.simulation}b, ~\ref{f.simulation}c, and ~\ref{f.simulation}d, respectively.

\section{Methodology}
\label{s.methodology}

This section provides a brief description of the datasets employed in this work, i.e., CIFAR-10 and CIFAR-100, as well all the setup considered during the experiments.

\subsection{Dataset}
\label{ss.dataset}

In this work, we consider the public datasets CIFAR-10 and CIFAR-100~\cite{cifar} to evaluate the performance of MaxDropoutV2 since both datasets are widely employed in similar regularization contexts~\cite{zhong2020random,cutout,maxdropout,localdrop}. Both datasets comprise $60,000$ color images of animals, automobiles, and ships, to cite a few, with a size of $32\times32$ pixels. Such images are divided such that $50,000$ instances are employed for training, and $10,000$ samples are considered for evaluation purposes. The main difference between CIFAR-10 and CIFAR-100 regards the number of classes, i.e., CIFAR-10 comprises $10$ classes while CIFAR-100 is composed of $100$ classes. 


\subsection{Experimental Setup}
\label{ss.experimental}

To provide a fair comparison, we adopted the same protocol employed in several works in literature~\cite{cutout,maxdropout,zhong2020random}, which evaluate the proposed techniques over the ResNet-18~\cite{resnet} neural network. Regarding the pre-processing steps, each image sample is resized to $32\times32$ pixels for further extracting random crops of size $28\times28$ pixels, with the addition of horizontal flip. The network hyperparameter setup employs the Stochastic Gradient Descent (SGD) with Nesterov momentum of $0.9$ and a weight decay of \num{5e-4}. The initial learning rate is initially set to $0.1$ and updated on epochs $60$, $120$, and $160$ by multiplying its value by $0.2$. Finally, the training is performed during a total of $200$ epochs and repeated during five rounds over each dataset to extract statistical measures. It is important to highlight that this protocol is used in several other works related to regularization on Deep Learning models~\cite{cutout,maxdropout,randomerasing}. In this work, we compare our proposed method against other regularizers that explicitly target to improve the results of CNNs.

Regarding the hardware setup, experiments were conducted using an Intel 2x Xeon\textregistered E5-2620 @ 2.20GHz with 40 cores, a GTX $1080$ Ti GPU, and $128$ GB of RAM.~\footnote{The code will be available in case of the paper acceptance.}

\section{Experimental Results}
\label{s.experiments}

This section provides an extensive set of experiments where MaxDropoutV2 is compared against several baselines considering both classification error rate and time efficiency. Additionally, it also evaluates combining MaxDropoutV2 with other regularization techniques.

\subsection{Classification Error}
\label{ss.classification}

Table~\ref{t.classificationError} shows the average error rate for all models and architectures regarding the task of image classification. Highlighted values denote the best results, which were obtained over five independent repetitions.

\begin{table}[!h]
    \centering
    \caption{Average classification error rate ($\%$) over CIFAR-10 and CIFAR-100 datasets. Notice ResNet18 results are provided as the baseline. }
    \label{t.classificationError}
    \begin{tabular}{lcc}
        \toprule
        \textbf{} & \textbf{CIFAR-10} & \textbf{CIFAR-100}
        \\ \midrule
        ResNet-18~\cite{resnet} & $4.72 \pm 0.21$ & $22.46 \pm 0.31$
        \\ \midrule
        Cutout~\cite{cutout} & $\bm{3.99 \pm 0.13}$ & $21.96 \pm 0.24$
        \\ \midrule
        RandomErasing~\cite{randomerasing} & $4.31 \pm 0.07$ & $24.03 \pm 0.19$
        \\ \midrule
        LocalDrop~\cite{localdrop} & $4.3$ & $22.2$
        \\ \midrule
        MaxDropout~\cite{maxdropout} & $4.66 \pm 0.13$ & $21.93 \pm 0.07$
        \\ \midrule
        MaxDropoutV2 (ours) & $4.63 \pm 0.04$ & $\bm{21.92 \pm 0.23}$
        \\ \bottomrule
    \end{tabular}
\end{table}

From the results presented in Table~\ref{t.classificationError}, one can observe that Cutout obtained the lowest error rate over the CIFAR-10 dataset. Meanwhile, MaxDropoutV2 achieved the most accurate results considering the CIFAR-100 dataset, showing itself capable of outperforming its first version, i.e., MaxDropout, over more challenging tasks composed of a higher number of classes.

\begin{figure*}[!ht]
\center

\begin{tabular}{cc}
	\includegraphics[width=0.51\textwidth]{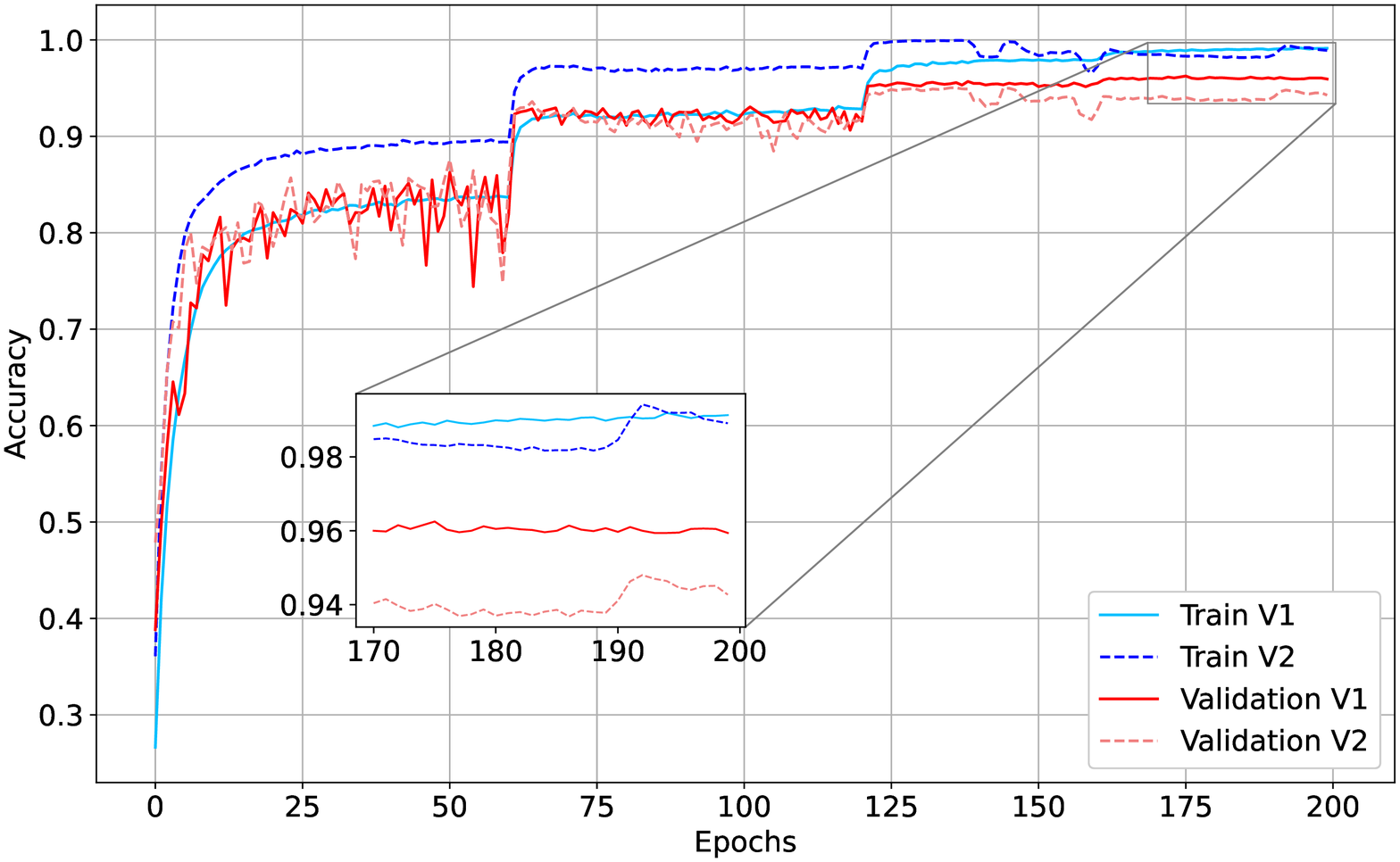} &
	\includegraphics[width=0.51\textwidth]{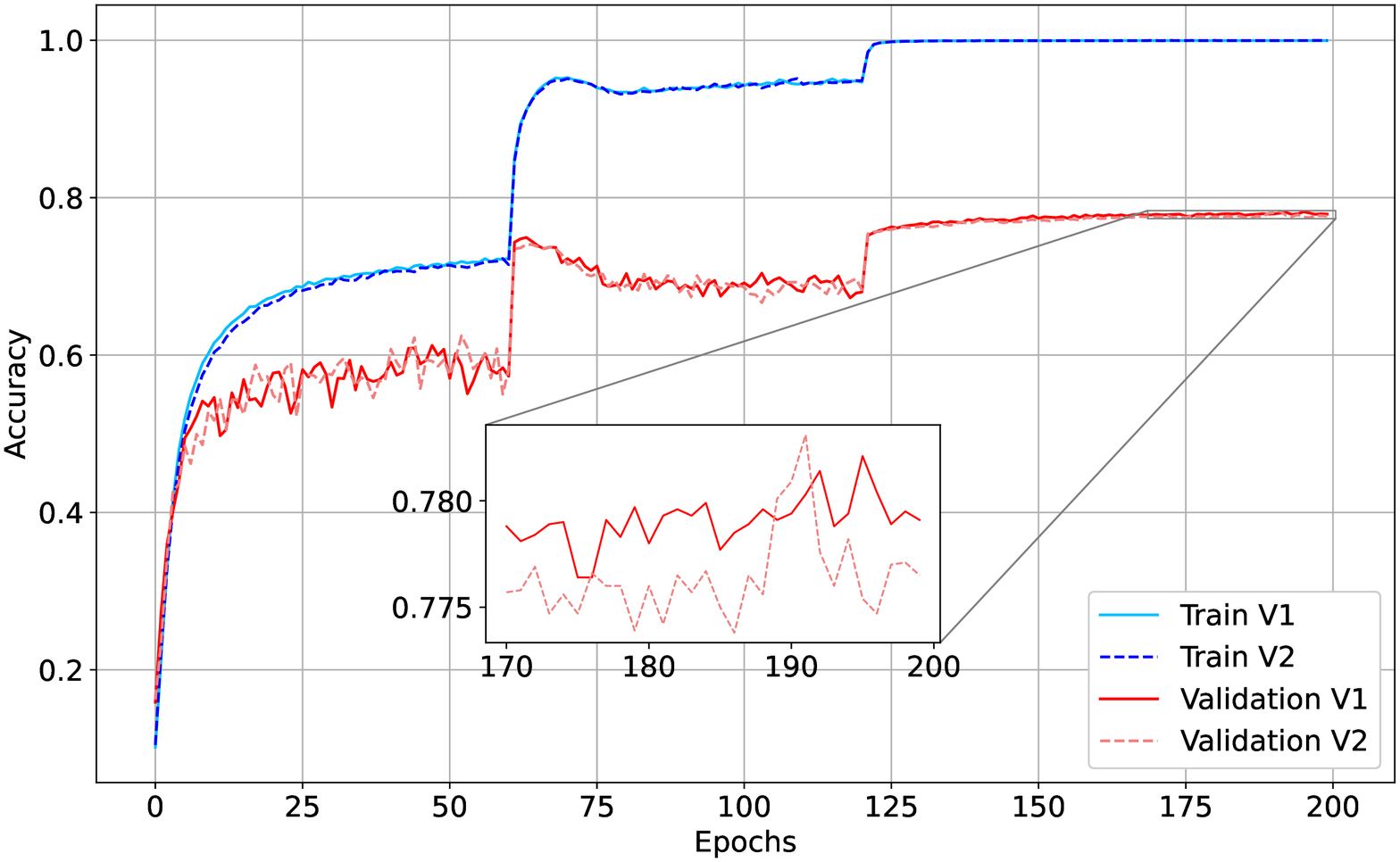}\\
\end{tabular}

\caption{Convergence analysis regarding the CIFAR-10 and CIFAR-100 datasets.}
	
\label{f.res_cifar}
\end{figure*}

Additionally, Figure~\ref{f.res_cifar} depicts the convergence evolution of MaxDropout and MaxDropoutV2 over the training and validation splits, in which the training partition comprises $50,000$ samples, and the validation contains $10,000$ samples. In Figure~\ref{f.res_cifar}, V1 stands for the MaxDropout method, and V2 stands for the proposed approach. One can notice that MaxDropoutV2 does not overpass the MaxDropout validation accuracy, mainly on the CIFAR-10 dataset. However, when dealing with more classes and the same number of training samples, both performances was almost the same, indicating the robustness of MaxDropoutV2.

\subsection{Combining Regularization Techiniques}
\label{ss.combiningTechinques}

A critical point about regularization concerns avoiding overfitting and improving the results of a given neural network architecture in any case. For instance, if some regularization approach is already applied, including another regularization should still improve the outcomes. In this context, MaxDropoutV2 performs this task with success, as shown in Table~\ref{t.accuracy.along}.

\begin{table}[!h]
    \centering
    \caption{Average classification error rate ($\%$) over CIFAR-10 and CIFAR-100 datasets combining MaxDropout and MaxDropoutV2 with Cutout.}
    \label{t.accuracy.along}
    \begin{tabular}{lcc}
        \toprule
        \textbf{} & \textbf{CIFAR-10} & \textbf{CIFAR-100}
        \\ \midrule
        ResNet-18~\cite{resnet} & $4.72 \pm 0.21$ & $22.46 \pm 0.31$
        \\ \midrule
        Cutout~\cite{cutout} & $3.99 \pm 0.13$ & $21.96 \pm 0.24$
        \\ \midrule
        MaxDropout~\cite{maxdropout} & $4.66 \pm 0.13$ & $21.93 \pm 0.07$
        \\ \midrule
        MaxDropoutV2 (ours) & $4.63 \pm 0.04$ & $21.92 \pm 0.23$
        \\ \midrule
        MaxDropout~\cite{maxdropout} + Cutout~\cite{cutout} & $\bm{3.76 \pm 0.08}$ & $\bm{21.82 \pm 0.13}$
        \\ \midrule
        MaxDropoutV2 (ours) + Cutout~\cite{cutout} & $3.95 \pm 0.13$ & $\bm{21.82 \pm 0.12}$
        \\ \bottomrule
    \end{tabular}
\end{table}

\subsection{Performance Evaluation}
\label{ss.performanceEvaluation}

The main advantage of MaxDropoutV2 over MaxDropout regards its computational time. In this context, Tables~\ref{t.time.cifar10} and~\ref{t.time.cifar100} provides the average time demanded to train both models considering each epoch and the total consumed time. Such results confirm the hypothesis stated in Section~\ref{ss.proposed} since MaxDropoutV2 performed around $10\%$ faster than the standard version.

\begin{table}[!ht]
    \centering
    \caption{Time consumed in seconds for training ResNet-18 in CIFAR-10 dataset.}
    \label{t.time.cifar10}
    \begin{tabular}{lcc}
        \toprule
        \textbf{} & \textbf{Seconds per Epoch} & \textbf{Total time}
        \\ \midrule
        MaxDropout~\cite{maxdropout} & $32.8$ & $6,563$
        \\ \midrule
        MaxDropoutV2 (ours) & $\bm{29.8}$ & $\bm{5,960}$
        \\ \bottomrule
    \end{tabular}
\end{table}

\begin{table}[!ht]
    \centering
    \caption{Time consumed in seconds for training ResNet-18 in CIFAR-100 dataset.}
    \label{t.time.cifar100}
    \begin{tabular}{lcc}
        \toprule
        \textbf{} & \textbf{Seconds per Epoch} & \textbf{Total time}
        \\ \midrule
        MaxDropout~\cite{maxdropout} & $33.1$ & $6,621$
        \\ \midrule
        MaxDropoutV2 (ours) & $\bm{30.2}$ & $\bm{6,038}$
        \\ \bottomrule
    \end{tabular}
\end{table}

\subsection{Evaluating Distinct Drop Rate Scenarios}
\label{ss.ablation}

\begin{sloppypar}
This section provides an in-depth analysis of MaxDropoutV2 and MaxDropout~\cite{maxdropout} results considering a proper selection of the drop rate parameter. Tables~\ref{t.ablation.cifar10} and~\ref{t.ablation.cifar100} present the models' results while varying the drop rate from $5$\% to $50$\% considering CIFAR-10 and CIFAR-100 datasets, respectively.
\end{sloppypar}

\begin{table}[!ht]
    \centering
    \caption{Mean error (\%) concerning CIFAR-10 dataset.}
    \label{t.ablation.cifar10}
    \begin{tabular}{lcc}
    	\toprule
    	\textbf{Drop Rate} & \textbf{MaxDropoutV2} & \textbf{MaxDropout }
    	\\ \midrule
    	$5$ & $\bm{4.63 \pm 0.03}$ & $4.76 \pm 0.09$
    	\\ \midrule
    	$10$ & $4.67 \pm 0.13$ & $4.71 \pm 0.09$
    	\\ \midrule
    	$15$ & $4.76 \pm 0.12$ & $\bm{4.63 \pm 0.11}$
    	\\ \midrule
    	$20$ & $4.66 \pm 0.13$ & $4.70 \pm 0.08$
    	\\ \midrule
    	$25$ & $4.75 \pm 0.11$ & $4.70 \pm 0.06$
    	\\ \midrule
    	$30$ & $4.63 \pm 0.16$ & $4.67 \pm 0.12$
    	\\ \midrule
    	$35$ & $4.70 \pm 0.18$ & $4.71 \pm 0.16$
    	\\ \midrule
    	$40$ & $4.74 \pm 0.13$ & $4.79 \pm 0.20$
    	\\ \midrule
    	$45$ & $4.65 \pm 0.16$ & $4.71 \pm 0.11$
    	\\ \midrule
    	$50$ & $4.71 \pm 0.04$ & $4.75 \pm 0.10$
    	\\ \bottomrule
    \end{tabular}
\end{table}

Even though MaxDropoutV2 did not achieve the best results in Table~\ref{t.classificationError}, the results presented in Table~\ref{t.ablation.cifar10} show the technique is capable of yielding satisfactory outcomes considering small drop rate values, i.e., $5\%$, while the standard model obtained its best results considering a drop rate of $15\%$. Additionally, one can notice that MaxDropoutV2 outperformed MaxDropout in eight-out-of-ten scenarios, demonstrating the advantage of the model over distinct circumstances.

\begin{table}[!ht]
    \centering
    \caption{Mean error (\%) concerning CIFAR-100 dataset.}
    \label{t.ablation.cifar100}
    \begin{tabular}{lcc}
        \toprule
        \textbf{Drop Rate} & \textbf{MaxDropoutV2} & \textbf{MaxDropout }
        \\ \midrule
        $5$ & $22.26 \pm 0.31$ & $22.05 \pm 0.17$
        \\ \midrule
        $10$ & $22.19 \pm 0.13$ & $22.06 \pm 0.32$
        \\ \midrule
        $15$ & $22.25 \pm 0.23$ & $22.16 \pm 0.20$
        \\ \midrule
        $20$ & $22.26 \pm 0.30$ & $21.98 \pm 0.21$
        \\ \midrule
        $25$ & $22.02 \pm 0.13$ & $\bm{21.93 \pm 0.23}$
        \\ \midrule
        $30$ & $\bm{21.92 \pm 0.23}$ & $22.07 \pm 0.24$
        \\ \midrule
        $35$ & $22.00 \pm 0.07$ & $22.10 \pm 0.29$
        \\ \midrule
        $40$ & $22.09 \pm 0.16$ & $22.16 \pm 0.34$
        \\ \midrule
        $45$ & $21.95 \pm 0.15$ & $22.31 \pm 0.29$
        \\ \midrule
        $50$ & $22.13 \pm 0.19$ & $22.33 \pm 0.23$
        \\ \bottomrule
    \end{tabular}
\end{table}

In a similar fashion, Table~\ref{t.ablation.cifar100} provides the mean classification error considering distinct drop rate scenarios over CIFAR-100 dataset. In this context, both techniques required larger drop rates to obtain the best results, i.e., $25$ and $30$ for MaxDropout and MaxDropoutV2, respectively.  Moreover, MaxDropoutV2 outperformed MaxDropout in all cases when the drop rates are greater or equal to $30$, showing more complex problems demand higher drop rates.

\subsection{Discussion}
\label{ss.discussion}


According to the provided results, the proposed method accomplishes at least equivalent outcomes to the original MaxDropout, outperforming it in terms of classification error in most cases. Moreover, MaxDropoutV2 presented itself as a more efficient alternative, performing around $10\%$ faster than the previous version for the task of CNN training.

The main drawback regarding MaxDropoutV2 is that the model is cemented to the network architecture, while MaxDropout applicability is available to any network's architecture. In a nutshell, MaxDropoutV2 relies on a matrix or high dimensional tensors designed to accommodate CNNs' layers outputs, while the standard MaxDropout works well for any neural network structure, such as Multilayer Perceptrons and Transformers, for instance.

\section{Conclusion and Future Works}
\label{s.conclusion}

This paper presented an improved version of the regularization method MaxDropout, namely MaxDropoutV2, which stands for a tailored made regularization technique for convolutional neural networks. In short, the technique relies on a more representative feature space to accommodate the convolutional layer outputs.

Experimental results showed the method significantly reduced the time demanded to train the network, performing around $10\%$ faster than the standard MaxDropout with similar or more accurate results. Moreover, it demonstrated that MaxDropoutV2 is more robust to the selection of the drop rate parameter. Regarding future work, we will evaluate MaxDropoutV2 in distinct contexts and applications, such as object detection and image denoising.

%
%
%
\bibliographystyle{splncs04}
\bibliography{refs}

\begin{thebibliography}{10}
\providecommand{\url}[1]{\texttt{#1}}
\providecommand{\urlprefix}{URL }
\providecommand{\doi}[1]{https://doi.org/#1}

\bibitem{cutout}
DeVries, T., Taylor, G.W.: Improved regularization of convolutional neural
  networks with cutout. arXiv preprint arXiv:1708.04552  (2017)

\bibitem{santos2021Covid}
{dos Santos}, C.F.G., Passos, L.A., {de Santana}, M.C., Papa, J.P.: Normalizing
  images is good to improve computer-assisted covid-19 diagnosis. In: Kose, U.,
  Gupta, D., {de Albuquerque}, V.H.C., Khanna, A. (eds.) Data Science for
  COVID-19, pp. 51--62. Academic Press (2021).
  \doi{https://doi.org/10.1016/B978-0-12-824536-1.00033-2},
  \url{https://www.sciencedirect.com/science/article/pii/B9780128245361000332}

\bibitem{gal2017concrete}
Gal, Y., Hron, J., Kendall, A.: Concrete dropout. In: Advances in Neural
  Information Processing Systems. pp. 3581--3590 (2017)

\bibitem{resnet}
He, K., Zhang, X., Ren, S., Sun, J.: Deep residual learning for image
  recognition. In: Proceedings of the IEEE conference on computer vision and
  pattern recognition. pp. 770--778 (2016)

\bibitem{ioffe2015batch}
Ioffe, S., Szegedy, C.: Batch normalization: Accelerating deep network training
  by reducing internal covariate shift. arXiv preprint arXiv:1502.03167  (2015)

\bibitem{kingma2015variational}
Kingma, D.P., Salimans, T., Welling, M.: Variational dropout and the local
  reparameterization trick. In: Advances in Neural Information Processing
  Systems. pp. 2575--2583 (2015)

\bibitem{cifar}
Krizhevsky, A., Nair, V., Hinton, G.: Cifar-10 and cifar-100 datasets. URl:
  https://www. cs. toronto. edu/kriz/cifar. html  \textbf{6}, ~1 (2009)

\bibitem{localdrop}
Lu, Z., Xu, C., Du, B., Ishida, T., Zhang, L., Sugiyama, M.: Localdrop: A
  hybrid regularization for deep neural networks. IEEE Transactions on Pattern
  Analysis and Machine Intelligence  (2021)

\bibitem{molchanov2017variational}
Molchanov, D., Ashukha, A., Vetrov, D.: Variational dropout sparsifies deep
  neural networks. In: Proceedings of the 34th International Conference on
  Machine Learning-Volume 70. pp. 2498--2507. JMLR. org (2017)

\bibitem{noda2015audio}
Noda, K., Yamaguchi, Y., Nakadai, K., Okuno, H.G., Ogata, T.: Audio-visual
  speech recognition using deep learning. Applied Intelligence  \textbf{42}(4),
   722--737 (2015)

\bibitem{passosECCOMAS19}
Passos, L.A., Santos, C., Pereira, C.R., Afonso, L.C.S., Papa, J.P.: A hybrid
  approach for breast mass categorization. In: ECCOMAS Thematic Conference on
  Computational Vision and Medical Image Processing. pp. 159--168. Springer
  (2019)

\bibitem{RoderICAISC:20}
Roder, M., Passos, L.A., Ribeiro, L.C.F., Benato, B.C., Falc\~{a}o, A.L., Papa,
  J.P.: Intestinal parasites classification using deep belief networks. In: The
  19th International Conference on Artificial Intelligence and Soft Computing
  (ICAISC). IEEE (In Press)

\bibitem{Roder:20EDrop}
Roder, M., de~Rosa, G.H., de~Albuquerque, V.H.C., Rossi, A.e.L.D., Papa,
  J.a.P.: Energy-based dropout in restricted boltzmann machines: Why not go
  random. IEEE Transactions on Emerging Topics in Computational Intelligence
  pp. 1--11 (2020). \doi{10.1109/TETCI.2020.3043764}

\bibitem{santanaIEEE-IS:19}
Santana, M.C., Passos, L.A., Moreira, T.P., Colombo, D., de~Albuquerque,
  V.H.C., Papa, J.P.: A novel siamese-based approach for scene change detection
  with applications to obstructed routes in hazardous environments. IEEE
  Intelligent Systems  \textbf{35}(1),  44--53 (2019)

\bibitem{maxdropout}
Santos, C.F.G.d., Colombo, D., Roder, M., Papa, J.P.: Maxdropout: Deep neural
  network regularization based on maximum output values. In: Proceedings of
  25th International Conference on Pattern Recognition, {ICPR} 2020, Milan,
  Italy, 10-15 January, 2021. pp. 2671--2676. {IEEE} Computer Society (2020)

\bibitem{simon2016imagenet}
Simon, M., Rodner, E., Denzler, J.: Imagenet pre-trained models with batch
  normalization. arXiv preprint arXiv:1612.01452  (2016)

\bibitem{srivastava2014dropout}
Srivastava, N., Hinton, G., Krizhevsky, A., Sutskever, I., Salakhutdinov, R.:
  Dropout: a simple way to prevent neural networks from overfitting. The
  journal of machine learning research  \textbf{15}(1),  1929--1958 (2014)

\bibitem{dropout}
Srivastava, N., Hinton, G., Krizhevsky, A., Sutskever, I., Salakhutdinov, R.:
  Dropout: a simple way to prevent neural networks from overfitting. The
  journal of machine learning research  \textbf{15}(1),  1929--1958 (2014)

\bibitem{sun2019idiopathic}
Sun, Z., He, S.: Idiopathic interstitial pneumonias medical image detection
  using deep learning techniques: A survey. In: Proceedings of the 2019 ACM
  Southeast Conference. pp. 10--15 (2019)

\bibitem{tan2019efficientnet}
Tan, M., Le, Q.V.: Efficientnet: Rethinking model scaling for convolutional
  neural networks. arXiv preprint arXiv:1905.11946  (2019)

\bibitem{wang2017gated}
Wang, J., Hu, X.: Gated recurrent convolution neural network for {OCR}. In:
  Advances in Neural Information Processing Systems. pp. 335--344 (2017)

\bibitem{wang2013fast}
Wang, S., Manning, C.: Fast dropout training. In: international conference on
  machine learning. pp. 118--126 (2013)

\bibitem{zhang2017beyond}
Zhang, K., Zuo, W., Chen, Y., Meng, D., Zhang, L.: Beyond a gaussian denoiser:
  Residual learning of deep cnn for image denoising. IEEE Transactions on Image
  Processing  \textbf{26}(7),  3142--3155 (2017)

\bibitem{zhong2020random}
Zhong, Z., Zheng, L., Kang, G., Li, S., Yang, Y.: Random erasing data
  augmentation. In: Proceedings of the AAAI Conference on Artificial
  Intelligence (AAAI) (2020)

\bibitem{randomerasing}
Zhong, Z., Zheng, L., Kang, G., Li, S., Yang, Y.: Random erasing data
  augmentation. In: AAAI. pp. 13001--13008 (2020)

\end{thebibliography}

%




\end{document}